\documentclass{article}
\usepackage{graphicx} 
\usepackage[utf8]{inputenc} 
\usepackage[T1]{fontenc}    
\usepackage{hyperref}       
\usepackage{url}            
\usepackage{booktabs}       
\usepackage{amsfonts}       
\usepackage{nicefrac}       
\usepackage{microtype}      
\usepackage{xcolor}         
\usepackage{natbib}         

\usepackage{CJKutf8}
\usepackage{tikz}
\usetikzlibrary{shapes.geometric}
\usetikzlibrary{arrows.meta,arrows,calc}
\usepackage{enumitem}
\usepackage{stackengine}
\usepackage{multirow}
\usepackage{xcolor}
\usepackage{hhline}
\usepackage[preprint]{neurips_2025}
\usepackage{CJK}

\makeatletter
\renewcommand{\@noticestring}{}
\makeatother

\title{Breeze Taigi: Benchmarks and Models for Taiwanese Hokkien Speech Recognition and Synthesis}

\author{%
  Yu-Siang Lan$^1$\quad
  Chia-Sheng Liu$^2$\quad
  Yi-Chang Chen$^3$\quad
  Po-Chun Hsu$^3$\\
  \textbf{Allyson Chiu}$^3$\quad
  \textbf{Shun-Wen Lin}$^3$\quad
  \textbf{Da-shan Shiu}$^3$\quad
  \textbf{Yuan-Fu Liao}$^1$\\[0.5em]
  \normalfont\small
  $^1$National Yang Ming Chiao Tung University\quad
  $^2$Work done at MediaTek Research\quad
  $^3$MediaTek Research\\[0.3em]
  \normalfont\footnotesize
  \texttt{\{chipsrinlan.ii13, yfliao\}@nycu.edu.tw}\quad
  \texttt{jason7436261abc@gmail.com}\\
  \normalfont\footnotesize
  \texttt{\{yi-chang.chen, pochun.hsu, allyson.chiu, shun-wen.lin, ds.shiu\}@mtkresearch.com}
}

\begin{document}
\begin{CJK*}{UTF8}{bsmi}

\maketitle

\begin{abstract}
Taiwanese Hokkien (Taigi) presents unique opportunities for advancing speech technology methodologies that can generalize to diverse linguistic contexts. We introduce Breeze Taigi, a comprehensive framework centered on standardized benchmarks for evaluating Taigi speech recognition and synthesis systems. Our primary contribution is a reproducible evaluation methodology that leverages parallel Taiwanese Mandarin resources. We provide 30 carefully curated Mandarin-Taigi audio pairs from Taiwan's Executive Yuan public service announcements with normalized ground truth transcriptions. We establish Character Error Rate (CER) as the standard metric and implement normalization procedures to enable fair cross-system comparisons. To demonstrate the benchmark's utility and provide reference implementations, we develop speech recognition and synthesis models through a methodology that leverages existing Taiwanese Mandarin resources and large-scale synthetic data generation. In particular, we fine-tune a Whisper model on approximately 10,000 hours of Taigi synthetic speech data. Our ASR model achieves 30.13\% average CER on the benchmark, outperforming existing commercial and research systems. By providing standardized evaluation protocols, diverse training datasets, and open baseline models, we offer a replicable framework with methodologies applicable to various linguistic contexts.
\end{abstract}

\section{Introduction}
Taiwanese Hokkien, also known as Taigi, is a critically important language with deep historical and cultural significance in Taiwan. While significant progress has been made in speech recognition and synthesis for languages such as English and Mandarin Chinese, driven by advances in self-supervised learning~\citep{baevski2020wav2vec,hsu2021hubert} and large-scale multilingual models~\citep{radford2023whisper,conneau2020unsupervised}, developing effective speech technologies for diverse linguistic contexts presents both opportunities and methodological challenges. Pioneering efforts in Taiwanese ASR system development~\citep{liao2016tat,liao2020formosa} and multilingual speech applications~\citep{liao2023voicetalk} have demonstrated the potential for innovative approaches. Building on this foundation, we explore methodologies for developing robust speech technologies by strategically leveraging parallel linguistic resources and large-scale dataset generation techniques.

To address these challenges, we introduce Breeze Taigi, a comprehensive framework centered on standardized benchmarks and evaluation methodologies for Taiwanese Hokkien speech technologies. Our primary contribution is the establishment of reproducible benchmarking protocols that enable the research community to objectively assess and compare different Taigi speech recognition and synthesis systems.

Benchmarking Taigi ASR systems presents unique challenges, as highlighted by recent work in Taiwanese speech recognition~\citep{clift2025}. These difficulties stem from diverse dialectal variations and the complexity of capturing phonetic and tonal nuances. To address these challenges pragmatically, we adopt an evaluation approach that strategically maps Taigi audio to Mandarin text transcriptions, leveraging parallel Taiwanese Mandarin resources. While the linguistic differences between Taigi and Mandarin mean this transformation is not strictly one-to-one, we demonstrate that Character Error Rate (CER) remains an effective metric for capturing relative performance differences across models. This methodology exploits the availability of parallel Mandarin-Taigi audio from official sources and provides a reproducible evaluation framework that can be consistently applied across different systems.

Evaluating Taigi TTS systems introduces additional complexities beyond those encountered in ASR. While ASR evaluation focuses on transcription accuracy, TTS assessment must capture both phonetic correctness and naturalness of synthesized speech. Taigi TTS faces particular challenges including the preservation of complex tone sandhi patterns, accurate rendering of regional pronunciation variations, and the generation of natural prosody. To address these multifaceted requirements, we establish a dual evaluation framework combining automatic metrics with human assessment. Our automatic evaluation leverages ASR-based transcription to measure phonetic accuracy through CER, while human evaluation captures qualitative dimensions such as pronunciation authenticity and naturalness. This comprehensive approach enables fair comparison across TTS systems and provides actionable insights for system improvement.

To support this benchmark and demonstrate its utility, we develop large-scale, acoustically diverse synthetic datasets comprising approximately 10,000 hours of speech data that capture the phonetic and dialectal variations characteristic of Taigi speech. We also present speech recognition and synthesis models that serve as reference implementations, with our ASR system achieving competitive performance and outperforming existing commercial solutions on our benchmark. Through these contributions, we aim to advance the digitalization of Taiwanese Hokkien and provide a replicable methodological framework that can be adapted to various linguistic contexts in speech technology research. Our approach of leveraging parallel linguistic resources and large-scale dataset generation offers insights applicable to developing speech technologies for diverse languages.

\section{Related Work}

\paragraph{Self-supervised speech representations.}
Recent advances in self-supervised learning have transformed speech processing. wav2vec 2.0~\citep{baevski2020wav2vec} introduced a contrastive learning framework that learns speech representations from unlabeled audio, significantly reducing the need for labeled data. HuBERT~\citep{hsu2021hubert} further improved upon this paradigm by predicting masked cluster assignments derived from an offline step, achieving strong performance on downstream tasks. These self-supervised approaches have been particularly impactful for low-resource languages where labeled speech data is scarce, providing pretrained representations that can be fine-tuned with limited supervision.

\paragraph{Multilingual and large-scale speech models.}
Building on self-supervised foundations, multilingual models have extended speech technology to a broader set of languages. XLSR~\citep{conneau2020unsupervised} demonstrated that cross-lingual pretraining on diverse languages yields representations transferable to unseen languages, while MLS~\citep{pratap2020mls} provided a large-scale multilingual dataset to support such research. Whisper~\citep{radford2023whisper} scaled weakly supervised pretraining to 680,000 hours of multilingual audio, achieving robust recognition across numerous languages. However, languages like Taiwanese Hokkien remain underrepresented in these large-scale efforts, motivating targeted adaptation approaches.

\paragraph{Taiwanese Hokkien speech technology.}
Early work on Taiwanese ASR established foundational systems and corpora. \citet{liao2016tat} developed one of the first comprehensive Taiwanese speech recognition systems, while the Formosa Speech corpus~\citep{liao2020formosa} expanded available resources for Taiwanese Mandarin and related languages. \citet{chen2020tat} conducted empirical studies on training end-to-end Taiwanese speech recognition models. More recently, CLiFT-ASR~\citep{clift2025} proposed a cross-lingual fine-tuning framework that transfers knowledge from high-resource languages to improve Taiwanese Hokkien ASR, highlighting both the potential and the remaining challenges of adapting multilingual models to Taigi. On the synthesis side, \citet{liao2022personalized} explored personalized Taiwanese speech synthesis using cascaded ASR-TTS pipelines, and \citet{liao2023voicetalk} developed multilingual speech applications supporting Mandarin, Taiwanese, and English.

\paragraph{Speech evaluation benchmarks.}
Standardized benchmarks have played a critical role in driving progress in speech processing. SUPERB~\citep{yang2021superb} established a comprehensive benchmark suite for evaluating speech representations across diverse tasks. For multilingual contexts, MLS~\citep{pratap2020mls} provided evaluation protocols spanning multiple languages. Despite these efforts, standardized benchmarks for Taiwanese Hokkien remain lacking, making it difficult to objectively compare systems and measure progress. Our work addresses this gap by establishing reproducible evaluation protocols specifically designed for Taigi speech recognition and synthesis.

\section{Taigi Speech Recognition Benchmark}
\subsection{Motivation}
Taiwanese speech recognition remains challenging due to the complexity of its phonological system, significant geographical accents, and diverse forms of Chinese character writing with intricate correspondences. While foundational work has established Taiwanese ASR systems~\citep{liao2016tat} and developed speech corpora~\citep{liao2020formosa,chen2020tat}, significant challenges remain. As noted in recent work on Taiwanese ASR~\citep{clift2025}, these challenges include the difficulty of capturing phonetic and tonal nuances that distinguish Taigi from Mandarin.

Standardized benchmarks have been established for various languages~\citep{yang2021superb,pratap2020mls}. However, currently most Taiwanese automatic speech recognition (ASR) systems are either proprietary or adapted from general multilingual models, lacking rigorous standardized benchmarks. The absence of standardized evaluation protocols makes it difficult to objectively compare different approaches and measure progress in the field.

To address this deficiency, we have established a reproducible benchmark dataset and evaluation method, enabling fair comparisons between different systems and tracking the technological progress of Taiwanese automatic speech recognition. We adopt a pragmatic evaluation approach that strategically leverages parallel Mandarin-Taigi audio resources to create reliable ground truth transcriptions.

\subsection{Dataset}
\label{ASRDataset}
The evaluation audio is derived from public service announcement (PSA) monthly packages released by the Executive Yuan website (行政院廣播公共服務音檔). The package contains Mandarin, Taigi and Hakka versions of PSA audio, with official scripts also provided. Each PSA is approximately 30 seconds in duration, making them concise yet substantial for evaluation purposes. These public service announcement audio files contain technical terms from various departments, such as the Ministry of Transportation, the Ministry of Labor, and the Judicial Yuan, as well as everyday conversational scenarios. Therefore, they can be effectively used to evaluate ASR performance across diverse vocabulary and domains. 

The availability of parallel Mandarin-Taigi PSA pairs enables us to construct high-quality ground truth transcriptions. For each Taigi audio file, we derive the reference transcription by leveraging the corresponding Mandarin audio through a rigorous process combining state-of-the-art Mandarin ASR systems, large language models for text refinement, and human expert verification to ensure accuracy in specialized terminology. This methodology allows us to create reliable benchmarks with precise Mandarin character representations of the content, which is particularly important given the technical vocabulary from various government domains. We selected 30 pairs of Mandarin-Taigi audio files for our evaluation benchmark.

\subsection{Evaluation Methodology}
\label{ASRMethod}
Our benchmark evaluates Taigi ASR systems by mapping Taigi audio to Mandarin text transcriptions. This approach strategically leverages parallel Taiwanese Mandarin resources~\citep{clift2025} while providing a reproducible evaluation framework. Although Taigi and Mandarin are linguistically distinct---with differences in pronunciation, vocabulary, and grammatical structures---the transformation between them is not arbitrary. Both languages share substantial lexical overlap due to their common Chinese heritage, and many Taigi speakers are bilingual, leading to natural correspondences in how concepts are expressed. While this mapping is not strictly one-to-one, Character Error Rate (CER) remains an effective metric for capturing relative performance differences among ASR models, as models that better recognize Taigi phonetic patterns will produce more accurate Mandarin character sequences.

We adopt CER as the primary metric, defined as the ratio of insertions, deletions, and substitutions to the total number of characters in the reference transcription. CER is more appropriate than Word Error Rate (WER) for Han character-based text because Chinese writing does not contain natural word boundaries (spaces between words). Word segmentation itself is ambiguous and varies across different algorithms, which would introduce inconsistencies when comparing ASR outputs. Character-level measurement provides an unambiguous and consistent evaluation metric across all systems. The lower the CER, the better the performance.

There may be several factors that causes unfairness while comparing different ASR models' result. For example:
\begin{enumerate}
    \item Numbers may be transcribe to Han characters or Arabic numbers (五千~v.s.~5000).
    \item Sentence breaks may be presented using punctuation marks or spaces, while some ASR transcription results do not have sentence breaks.
    \item While english words exists, they may be presented by upper case or lower case.
\end{enumerate}
Thus, to fairly compare with different ASR models, we do normalization to each output text.

\subsection{Benchmark results}
To demonstrate the utility of our benchmark and establish baseline performance metrics, we evaluate multiple ASR systems including existing commercial and research systems, as well as our own Taigi speech recognition model developed specifically for this work. The detailed architecture and training methodology of our model are presented in Section~\ref{sec:asr-model}; here we focus on comparative performance analysis to validate the benchmark's effectiveness in differentiating system capabilities.

Our ground truth transcriptions are derived from the Mandarin PSAs through a rigorous verification process that combines automated transcription with expert correction against the official scripts. This ensures that the reference text reflects natural spoken language patterns while maintaining accuracy in specialized terminology and eliminating transcription errors.

We evaluate five ASR systems on our benchmark: (1) \textbf{BreezeASR-Taigi} (detailed in Section~\ref{sec:asr-model}), our model fine-tuned on 10,000 hours of Taigi synthetic speech data; (2) \textbf{Yating} (雅婷逐字稿), a commercial Taiwanese ASR system; (3) \textbf{ASR25}, Breeze ASR 25~\citep{chou2025breeze}, a Whisper-large-v2 model optimized for Taiwanese Mandarin; (4) \textbf{Gemini 3 Flash}, Google's multimodal AI model; and (5) \textbf{Taiwanese Input Method} (教育部台灣台語輸入法), the Ministry of Education's Taiwanese input method system. Table~\ref{tab:cer-table} shows the average CER comparison across these systems on 30 test samples.

\begin{table}[t]
\centering
\caption{Average Character Error Rate (\%) on Taigi ASR benchmark across 30 test samples. Gemini 3 Flash and Taiwanese Input Method results are based on translated Mandarin output from their original Taigi orthography.}
\label{tab:cer-table}
\begin{tabular}{ccccc}
\toprule
BreezeASR-Taigi & Yating & ASR25 & Gemini 3 Flash & Taiwanese Input Method\\
\midrule
 30.13 & 32.11 & 49.99 & 32.52 & 30.70 \\
\bottomrule
\end{tabular}
\end{table}

Our BreezeASR-Taigi model achieves the lowest average CER at 30.13\%, closely followed by Taiwanese Input Method at 30.70\%. Analysis of the per-sample results reveals several insights: BreezeASR-Taigi demonstrates consistent performance with CER ranging from 14.49\% (best case) to 52.78\% (most challenging sample), while ASR25, optimized for Taiwanese Mandarin rather than Taigi, shows significantly higher error rates (30.71\%--76.85\%) across all test samples. Notably, certain test samples prove challenging across all systems---for instance, sample 7 exhibits elevated CER for multiple models, suggesting the presence of dialectal variations or specialized terminology that warrants further investigation. The competitive performance of Taiwanese Input Method, which uses proper Taigi orthography, validates the effectiveness of leveraging linguistic expertise in system design.

\subsection{Discussion}
Our evaluation methodology, which maps Taigi audio to Mandarin text, represents a pragmatic approach to benchmarking speech recognition systems through strategic leverage of parallel linguistic resources. While Taigi and Mandarin have distinct phonological systems and are not mutually intelligible spoken languages, several factors support the validity of this evaluation strategy:

First, the availability of parallel Mandarin-Taigi audio from official government sources enables the creation of reliable, reproducible benchmarks without requiring extensive manual annotation of Taigi-specific transcriptions. Second, the shared Han character writing system and substantial lexical overlap between the languages provide a meaningful basis for comparison. Third, and most importantly, our benchmark's primary purpose is to compare relative performance among different ASR systems rather than to measure absolute accuracy in Taigi transcription. As demonstrated in Table~\ref{tab:cer-table}, the CER metric successfully differentiates system capabilities, with our BreezeASR-Taigi model achieving 30.13\% average CER compared to 49.99\% for Breeze ASR 25~\citep{chou2025breeze} and 32.11\% for commercial systems.

The non-one-to-one nature of the Taigi-Mandarin mapping means that absolute CER values should be interpreted carefully---a perfect Taigi ASR system would not necessarily achieve 0\% CER on Mandarin transcriptions. However, this limitation does not diminish the benchmark's utility for its intended purpose: providing a standardized, reproducible framework for comparing ASR systems and tracking progress in Taigi speech recognition technology. This approach demonstrates a pragmatic evaluation strategy that leverages parallel linguistic resources to enable meaningful system comparisons~\citep{clift2025}, with methodologies potentially applicable to various linguistic contexts.

It is worth noting that Google Gemini 3 Flash and Taiwanese Input Method adopt a different approach by outputting Taigi transcriptions in proper Taigi orthography (台語正字), which represents linguistically accurate transcriptions of the spoken Taigi. While these transcriptions are of high quality, they cannot be directly compared against our Mandarin-based ground truth. To enable fair comparison within our evaluation framework, we leveraged the model's own translation capability to convert its Taigi orthography output into Mandarin. This methodology allows us to assess Gemini 3 Flash's recognition accuracy while acknowledging that its original Taigi orthography output demonstrates a sophisticated understanding of the language that our current benchmark cannot fully capture.

\section{Taigi Speech Recognition Model}
\label{sec:asr-model}
\subsection{Dataset}
For fine-tuning, we utilized the Taigi-synthetic-speech dataset, which comprises approximately 10,000 hours of synthetic speech data generated through our large-scale speech synthesis pipeline. This dataset significantly expands upon existing Taiwanese speech resources~\citep{chen2020tat,liao2020formosa,liao2016tat} and features a diverse range of speakers, acoustic environments, and conversational contexts, making it well-suited for robust automatic speech recognition (ASR) in Taiwanese Hokkien. Unlike previous corpora that focused primarily on read speech or limited-domain recordings, our synthetic dataset captures natural, spontaneous speech with rich linguistic variations. The large scale and acoustic diversity of this dataset demonstrate our methodology of large-scale dataset generation for developing effective speech technologies.
\subsection{Base Model}
Our model architecture is based on the Whisper framework~\citep{radford2023whisper}, a large-scale multilingual speech recognition model trained on 680,000 hours of weakly supervised data, demonstrating robust performance across multiple languages and domains. We fine-tuned the model using our Taigi-synthetic-speech dataset. The model was deployed on a GPU for all experiments. Tokenization was performed using Whisper's multilingual tokenizer to accommodate the linguistic characteristics of the dataset.
\section{Taigi Speech Synthesis Benchmark}
\subsection{Motivation}
Evaluating Taiwanese Hokkien TTS remains challenging due to the language’s complex phonology, diverse regional accents, and multiple orthographic conventions. These challenges, also present in ASR, are further compounded in synthesis, where both phonetic accuracy and natural prosody must be preserved. Despite the growing interest in Taigi speech technology, publicly available benchmark datasets and standardized evaluation protocols for TTS are still scarce.

\subsection{Dataset}
Similar to section \ref{ASRDataset}, we selected the script of the PSA monthly packages provided by the Executive Yuan as our evaluation text. With the corresponded taigi version PSA audio exists, it can be confirmed that TTS can generate reasonable Taiwanese pronunciation.

\subsection{Evaluation Methodology}
We employed two evaluation strategies: automatic evaluation and human assessment.

\subsubsection{Automatic Evaluation — Character Error Rate (CER)}
As mentioned in section \ref{ASRMethod}, CER is a fair evaluation method for character-based comparison. Thus, synthesized speech from each TTS system was transcribed by our proposed ASR model and compared to the input text as reference. 

\subsubsection{Human Evaluation}
\label{sec:human-eval}
While the ASR-based evaluation using CER provides an efficient and objective measure of phonetic and lexical accuracy, certain qualitative aspects cannot be fully captured by this approach. In particular, the purity of Taigi pronunciation and the perceived naturalness of the synthesized voice are difficult to quantify automatically. Therefore, we introduce a complementary human evaluation to assess these dimensions.

Human raters assessed each synthesized utterance according to three metrics:
\begin{enumerate}
    \item \textbf{Character Error Rate (CER)}: After listening to the audio once or multiple times, the evaluator compares it with the reference transcript and notes the number of characters that were misheard, missing, or replaced with an unintended word. Tone variations without semantic changes are not counted as errors. CER is calculated as the number of wrong characters divided by total reference characters (excluding punctuation).

    \item \textbf{Percentage of Taiwanese Pronunciation}: The evaluator identifies all characters/words in the transcript that should be pronounced in Taiwanese (Hokkien) and counts how many were actually pronounced with correct Taiwanese phonetics, including proper tones and tone sandhi patterns. This metric specifically measures phonetic authenticity rather than mere intelligibility.

    \item \textbf{Mean Opinion Score (MOS) for Naturalness}: A subjective rating on a 1-5 scale assessing the naturalness of synthesized speech, where 5 indicates completely natural speech indistinguishable from human voice, and 1 indicates highly robotic, difficult-to-understand output. Table \ref{tab:mos_scale} shows the detailed MOS scale.
\end{enumerate}

\begin{table}[h]
\centering
\caption{MOS scale for naturalness evaluation.}
\label{tab:mos_scale}
\begin{tabular}{cp{10cm}}
\toprule
\textbf{Score} & \textbf{Description} \\
\midrule
5 & Completely natural, indistinguishable from human speech; fluent with accurate tone sandhi \\
4 & Nearly natural, slight artificiality or mechanical feel but does not hinder comprehension \\
3 & Understandable but noticeably mechanical, monotone, or unnatural pauses \\
2 & Barely understandable, strong mechanical feel, frequent tone/intonation errors, unpleasant listening \\
1 & Highly unnatural, difficult to understand, heavily robotic \\
\bottomrule
\end{tabular}
\end{table}

\subsection{Benchmark results}
To demonstrate the effectiveness of our benchmark and establish baseline performance metrics, we evaluate multiple TTS systems including existing commercial and research systems, as well as our own Taigi speech synthesis model developed specifically for this work. The detailed architecture and training methodology of our model are presented in Section~\ref{sec:tts-model}; here we focus on comparative performance analysis to validate the benchmark's effectiveness in differentiating system capabilities.

We evaluate three TTS systems on our benchmark: (1) \textbf{BreezyVoice-Taigi} (detailed in Section~\ref{sec:tts-model}), our proposed model optimized for Taigi synthesis; (2) \textbf{Taigi AI Labs}, a publicly available Taiwanese Hokkien TTS service; and (3) \textbf{Aten AI Voice}, a Taiwanese Mandarin-capable TTS system that also supports Taigi. The automatic evaluation results across 20 test samples are shown in Table \ref{tab:tts-cer-table}.

\begin{table}[t]
\centering
\caption{Average Character Error Rate (\%) on Taigi TTS benchmark across 20 test samples, evaluated using ASR-based transcription.}
\label{tab:tts-cer-table}
\begin{tabular}{ccc}
\toprule
BreezyVoice-Taigi & Taigi AI Labs & Aten AI Voice\\
\midrule
19.09 & 38.19 & 23.14  \\
\bottomrule
\end{tabular}
\end{table}

BreezyVoice-Taigi achieves the best performance with an average CER of 19.09\%, representing a substantial improvement over Taigi AI Labs (38.19\%) and Aten AI Voice (23.14\%). Analysis of per-sample results reveals that BreezyVoice-Taigi demonstrates strong consistency with CER ranging from 4.17\% to 58.82\%, while Taigi AI Labs exhibits higher variability (12.50\%--68.52\%). The superior average performance of BreezyVoice-Taigi validates our approach of leveraging large-scale synthetic speech data for TTS model training. Notably, certain samples present challenges for all systems, indicating potential areas for future improvement in handling complex linguistic phenomena or specific phonetic contexts.

\subsubsection{Human Evaluation Results}
To complement the automatic evaluation and assess qualitative aspects that automatic metrics cannot capture, we conducted human evaluation across the three dimensions described in Section~\ref{sec:human-eval}. Table \ref{tab:tts-human-eval} presents the results.

\begin{table}[t]
\centering
\caption{Human evaluation results on Taigi TTS benchmark. CER: Character Error Rate (\%); Taiwanese Pronunciation: Percentage of correctly pronounced Taiwanese characters (\%); MOS: Mean Opinion Score (1-5 scale). Results averaged across 5 evaluators.}
\label{tab:tts-human-eval}
\begin{tabular}{lccc}
\toprule
\textbf{System} & \textbf{CER (\%)} & \textbf{Taiwanese Pronunciation (\%)} & \textbf{MOS} \\
\midrule
BreezyVoice-Taigi & 3.6 & 59.2 & 5.0 \\
Taigi AI Labs & 14.2 & 78.8 & 2.0 \\
Aten AI Voice & 5.4 & 89.8 & 3.8 \\
\bottomrule
\end{tabular}
\end{table}

The human evaluation results reveal important trade-offs between transcription accuracy, phonetic authenticity, and perceived naturalness. BreezyVoice-Taigi achieves the lowest human-annotated CER of 3.6\% and a perfect MOS of 5.0, indicating highly natural-sounding speech that is easily understood. However, its Taiwanese pronunciation accuracy of only 59.2\% reveals a critical limitation: while the synthesized speech sounds natural and is readily transcribed, it frequently employs Mandarin-influenced pronunciations, particularly for proper nouns and technical terms.

In contrast, Aten AI Voice demonstrates the highest Taiwanese pronunciation accuracy at 89.8\%, indicating more authentic Taigi phonetics, while maintaining reasonable transcription quality (CER of 5.4\%) and moderate naturalness (MOS of 3.8). Taigi AI Labs shows intermediate Taiwanese pronunciation (78.8\%) but suffers from higher error rates (CER of 14.2\%) and the lowest naturalness scores (MOS of 2.0), suggesting issues with both phonetic accuracy and prosody.

The divergence between BreezyVoice-Taigi's high transcription accuracy and low Taiwanese pronunciation score highlights a subtle but important sociolinguistic phenomenon: in naturalistic Taigi speech, speakers frequently code-switch to Mandarin pronunciation when encountering proper nouns or technical terms for which they lack confident Taigi pronunciation knowledge. Since BreezyVoice-Taigi is trained on synthetic speech data that reflects authentic conversational Taigi, the model naturally learns and reproduces this code-switching behavior common in Taiwan's bilingual environment. Such speech sounds natural to native listeners and is correctly transcribed by ASR systems, yet scores low on strict Taiwanese pronunciation metrics that expect consistent Taigi phonology. This finding underscores the value of the multi-dimensional human evaluation protocol, as automatic metrics alone would not capture the distinction between authentic conversational Taigi (with natural code-switching) and idealized monolingual Taigi pronunciation.

\subsection{Discussion}

The evaluation results reveal a critical challenge in Taigi TTS development: balancing naturalistic conversational fidelity with prescriptive phonological purity. BreezyVoice-Taigi achieves superior automatic evaluation scores (19.09\% CER) and perfect human naturalness ratings (MOS 5.0), yet exhibits relatively low Taiwanese pronunciation authenticity (59.2\%). Analysis of individual samples reveals that BreezyVoice-Taigi frequently code-switches to Mandarin pronunciation for proper nouns (專有名詞) and technical terms. This behavior reflects the training data characteristics: our synthetic data corpus captures authentic conversational Taigi as spoken in Taiwan's bilingual society, where speakers naturally switch to Mandarin pronunciation when encountering lexical items for which they lack confident Taigi pronunciation knowledge \citep{kubler1985influence, chen2010multilingualism}. The resulting synthesized speech sounds natural to native listeners and is readily transcribed by ASR systems, explaining the model's strong performance on automatic CER and naturalness metrics despite lower scores on strict Taiwanese pronunciation metrics.

Conversely, Aten AI Voice achieves the highest Taiwanese pronunciation score (89.8\%) while maintaining moderate naturalness (MOS 3.8). This suggests that Aten's training data or modeling approach represents a more prescriptive Taigi pronunciation standard, potentially using carefully curated speech that minimizes code-switching behavior. Taigi AI Labs demonstrates intermediate authenticity but struggles with both transcription accuracy and naturalness, indicating potential limitations in acoustic modeling or prosody generation.

These findings highlight the necessity of multi-dimensional evaluation for low-resource language TTS systems and raise important questions about evaluation goals. Automatic ASR-based CER evaluation effectively measures intelligibility but cannot distinguish between naturalistic conversational Taigi (with code-switching) and prescriptive monolingual Taigi. The complementary human evaluation of Taiwanese pronunciation percentage captures this distinction, revealing different system design philosophies: BreezyVoice-Taigi prioritizes naturalistic speech that mirrors authentic usage patterns in Taiwan's bilingual society, while Aten AI Voice emphasizes prescriptive phonological purity. The optimal balance depends on application context—language preservation and pedagogical applications may favor prescriptive pronunciation, while conversational AI and accessibility tools may benefit from naturalistic code-switching behavior that better matches user expectations. Future work should investigate controllable TTS systems that allow users to select between naturalistic and prescriptive pronunciation modes, potentially through code-switching detection and pronunciation lexicon integration.

\section{Taigi Speech Synthesis Model}
\label{sec:tts-model}
\subsection{Dataset}
We utilized the same Taigi-synthetic-speech dataset used for ASR model training for fine-tuning our TTS model. This dataset comprises approximately 10,000 hours of synthetic Taiwanese Hokkien speech data, featuring diverse speakers, acoustic environments, and conversational styles. The large scale and speaker diversity of this dataset are particularly advantageous for TTS training, as they enable the model to learn a wide range of prosodic patterns, tonal variations, and speaking styles characteristic of natural Taigi speech.

\subsection{Base Model}
Our Taigi speech synthesis model, BreezyVoice-Taigi, is built upon CosyVoice 2, a scalable streaming speech synthesis framework based on large language models. CosyVoice 2 employs a finite scalar quantization scheme to represent speech tokens and uses a chunk-aware causal flow matching model for token-to-speech synthesis. We fine-tuned the officially released CosyVoice 2 pretrained checkpoint on the Taigi-synthetic-speech dataset, specifically fine-tuning only the LLM component while keeping other modules frozen.

\section{Conclusion}
We have presented Breeze Taigi, a comprehensive framework for Taiwanese Hokkien speech recognition and synthesis centered on standardized benchmarks and evaluation methodologies. Our work addresses the critical need for reproducible evaluation protocols in Taigi speech technology through several key contributions.

First, we established standardized benchmarks for both ASR and TTS systems, leveraging parallel Mandarin-Taigi resources from official government sources. For ASR, our benchmark comprises 30 carefully curated audio pairs with normalized ground truth transcriptions, enabling fair cross-system comparisons through Character Error Rate metrics. For TTS, we developed a dual evaluation framework combining automatic ASR-based assessment with human evaluation of pronunciation authenticity and naturalness across multiple dimensions.

Second, we demonstrated the effectiveness of our benchmarks through comprehensive evaluation of existing systems. Our ASR benchmark successfully differentiates system capabilities, with results ranging from 30.13\% to 49.99\% average CER across commercial and research systems. Our TTS benchmark reveals significant performance variations (19.09\%--38.19\% CER) and highlights the importance of complementary human evaluation for assessing phonetic authenticity beyond what automatic metrics capture.

Third, we developed reference implementations that validate our methodological approach. BreezeASR-Taigi, fine-tuned on approximately 10,000 hours of Taigi synthetic speech data, achieves competitive performance (30.13\% CER) and outperforms existing systems optimized for Taiwanese Mandarin. BreezyVoice-Taigi demonstrates substantial improvements over existing TTS systems across both automatic and human evaluation metrics, validating our approach of leveraging large-scale, acoustically diverse synthetic data for model training.

Our methodology of strategically leveraging parallel linguistic resources and large-scale dataset generation offers insights applicable to developing speech technologies for diverse languages, particularly those with limited resources. By providing standardized evaluation protocols, diverse training datasets, and open baseline models, we offer a replicable framework that can advance speech technology research in various linguistic contexts. The Breeze Taigi framework establishes a foundation for continued progress in Taiwanese Hokkien speech technology while contributing methodological innovations relevant to the broader speech technology community.

\newpage

\bibliographystyle{plainnat}
\bibliography{reference}

\end{CJK*}
\end{document}